\documentclass[runningheads]{ola}

\usepackage{epsfig}

\begin{document}

\pagestyle{headings}

\mainmatter

\title{FFCNN: Fast FPGA based Acceleration for Convolution neural network inference}

\titlerunning{FFCNN: Fast FPGA based Acceleration for Convolution neural network inference}

\author{F. Keddous,\inst{1,2}, H-N. Nguyen\inst{2} \and A. Nakib\inst{1}}

\authorrunning{F. Keddous, H-N Nguyen and A. Nakib}

\institute{Universit\'e Paris Est Cr\'eteil, Laboratoire LISSI, 122 Rue P. Armangot, Vitry\\ Sur Seine, France \\
\email{nakib@u-pec.fr}
\and
 \\ Atos-BULL, Rue Jean Jaures, \\
 78340 Les Clayes-sous-Bois, France\\
\email{}
}

\maketitle
\section*{Abstract}
we present a new efficient OpenCL-based Accelerator for large scale Convolutional Neural Networks called “Fast Inference on FPGAs for Convolution Neural Network” (FFCNN). FFCNN is based on a deeply pipelined OpenCL kernels architecture. As pointed out before, high-level synthesis tools such as the OpenCL framework can easily port codes originally designed for CPUs/GPUs to FPGAs, but it is still difficult to make OpenCL codes run efficiently on FPGAs. This work aims to propose an efficient FPGA implementation of OpenCL High-Performance Computing Applications. To do so, a Data reuse and task mapping techniques are also presented to improve design efficiency.
In addition, the following motivations were taken into account when developing FFCNN:
• FFCNN has been designed to be easily implemented on Intel OpenCL SDK based FPGA design flow.
• In FFFCN, different techniques have been integrated to improve the memory band- with and throughput.
A performance analysis is conducted on two deep CNN for Large-Scale Images classifica- tion. The obtained results, and the comparison with other works designed to accelerate the same types of architectures, show the efficiency and the competitiveness of the pro- posed accelerator design by significantly improved performance and resource utilization.
\section{Introduction}
Recent work on the neural networks have shown great improvements over traditional machine learning algorithms. Especially in computer vision where a high adaptive capacity for a wide range of pattern recognition problems was demonstrated.
The convolutional neuron network (AlexNet)\cite{1} improved the classification accuracy of TOP-5 images in ImageNet \cite{2} datasets from 73.8\% to 84.7\% and helped to improve the performance of different computer vision problems \cite{3} with its ability to extract features. However, the complexity of its calculation and storage is high. According to current research, the size of the RN model continues to increase. In Table 1, we list the number of operations (additions or multiplications), the number of parameters and the top-1 precision on the ImageNet dataset \cite{2} of the Convolutional Neural Networks (CNN) models found in the literature for image classification, object detection, and image segmentation.

For instance, one of the largest and widely used CNN requires 39 billion floating point (FLOP) operations with an image size of $224\times 224$ and has a model parameter of 500 MB (VGG\cite{4}). The complexity of the calculations is proportional to the size of the input, then, the calculation of high resolution images will require more than 100 billion operations.

Therefore, it is important to select a computing architecture for any CNN based solution. A typical CPU runs 10 to 100 GFLOP per second. Energy efficiency is often less than 1 GOP per day. The CPUs are difficult to apply to cloud applications that require high performance in terms FLOP and mobile applications that require low power consumption. On the other hand, GPUs offer high performance up to 10 TOP per second. 

Usually, hardware accelerators are based on ASIC \cite{2} or FPGA \cite{3,4}. ASIC-based accelerators offer the highest performance and energy efficiency, but must withstand considerable development costs. Because of their reconfigurable nature, FPGA-based accelerators are more economical given development costs.

For years, FPGA developers have been struggling with difficult-to-use Register Transfer Level (RTL) programming languages such as VHDL and Verilog HDL. This makes programming a major issue for the FPGA. Thus, FPGA providers are beginning to provide high-level synthesis tools such as the OpenCL framework \cite{5} to enable FPGA programming using high-level languages.
Although developers can easily port codes originally designed for CPUs / GPUs to FPGAs with the OpenCL framework, it is still difficult to make OpenCL codes run efficiently on FPGAs. The same code may have different performance on different platforms because of the different execution methods related to the architecture. Therefore, developers must consider the FPGA architecture when optimizing OpenCL code.

The main contributions of this work are as follows: (1) an OpenCL based FPGA accelerator with an efficient pipelined kernel structure is proposed for large scale network (CNN) implementation; (2) the design space of the proposed architecture was fully explored on the Arria FPGA 10 and Stratix-10, two large-scale CNN models, were implemented and tested. The results show that the proposed scheme improves performance and resource utilization compared to previous work.

The rest of the paper is organized as follow: in the next section we recall CNN definition. In section 3, the proposed implementation is presented. The obtained results are shown in the section 4. The conclusion ends the paper.

\section{Convolution Neural Network}
 In this section, we present the basic functions of a neural network and we focus only on the inference procedure, which means that the Neural Network model was already trained and validated to predict or classify new data.
 


The basic architectural ideas of a Convolution Neural Network (CNN) \cite{CNNBasic} consist of the local receptive fields via the convolution operation and the spatial sub-sampling via the pooling operation. The Convolution operation can be formally written as:

\begin{equation}
	f_{x, y, h}^{C, l} = {\mathbf{w}_{h}^{l}}^{T} f_{x, y}^{O p, l-1}+b_{h}^{l}
\end{equation}

where $\mathbf{w}_{h}^{l}$ and $b_{h}^{l}$ are the weights and bias of the $h^{th}$ feature map, $f^{O p, l-1}$ and $f_{x, y, h}^{C, l}$ are the input and output feature maps, $l$ denotes the layer and $(x, y)$ is the spatial image coordinate. The superscript $C$ denotes convolution and $Op$ represents various operations, $e.g.$, input (when $l$ = 1), convolution, pooling, activation, etc.

Pooling applies local operations, $e.g.$, computing the maximum within a local neighborhood has the following form:

\begin{equation}
	f_{x, y, h}^{P_{max}, l} = max_{(m, n) \in \mathcal{N}_{x, y}} (f_{m, n, h}^{O p, l-1})
\end{equation}

where $\mathcal{N}_{x, y}$ denotes the local spatial neighborhood and $P_{max}$ denotes the max pooling. Often a spatial resolution reduction is applied after the max-pooling operation. Besides the two above-mentioned operations, there are several strategies applied within the CNN models, such as non-linear activation (e.g., the Rectified Linear Unit (ReLU) \cite{27}), dropout \cite{28} and batch normalization \cite{29}. A Fully Connected (FC) layer, can be added at the end of the concatenated layers. It takes all nodes (neurons) from the feature maps of the previous layer as input and connects it to every nodes (neurons) of the output feature map.
At the last layer, called dense layer, of the CNN models (referred to as the prediction layer), it is the common to use the Softmax activation function.

Then, the convolution (CONV) layers and the dense layer of fully connected layer (FC) layers are two common types of layers most of  architectures. CONV layers conduct two-dimensional (2D) convolutions on a set of input feature maps and add the results to get output feature maps. FC layers receive a feature vector as input and conduct matrix-vector multiplications. 

\begin{figure}
	\centering
	\begin{tabular}{cc}
		\includegraphics[height=1.5in, width=2in]{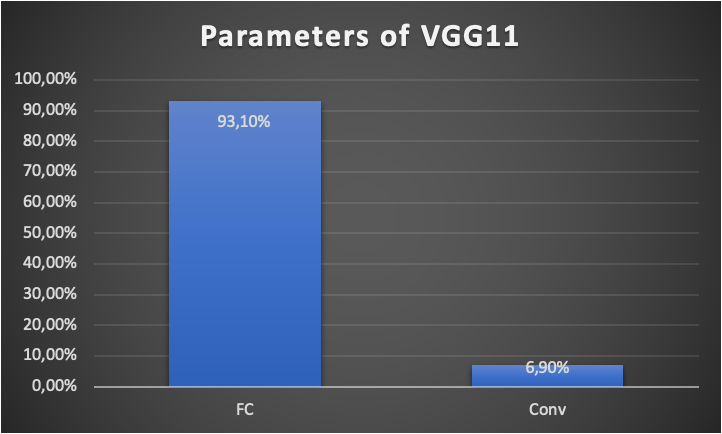} &
		\includegraphics[height=1.5in, width=2in]{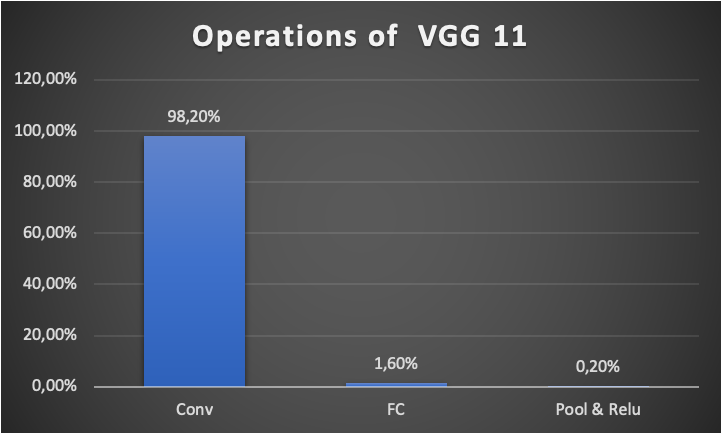}
	\\
	\end{tabular}   
	\caption{Distribution of the parameters and the operations in chain based architecture. Example of VGG with 11 layers.}
	\label{paramCount}
\end{figure}

Besides CONV and FC layers, NN layers also have pooling, ReLU, concat\cite{58}, elementwise\cite{22}, and other types of layers. But these layers contributes little to the computation and storage requirement of a neural network model. Figure\ref{paramCount} shows the distribution of weights and operations in the VGG-11 model. In this model, CONV and FC layers together contribute more than 99\% of the network’s weights and operations, which is similar to most of the CNN models. It is obvious that most of the neural network acceleration systems must be focus on these two types of layers.

\section{Proposed implementation}
In this work, we used an Altera FPGA Development Kit to build our CNN accelerator. In particular, the overall memory controller is a DDR3/DDR4 controller, the link controller is a PCIe controller, and the host computer is a desktop PC based on an x86 architecture.

\begin{figure}
	\centering
		\includegraphics[height=2.5in, width=4in]{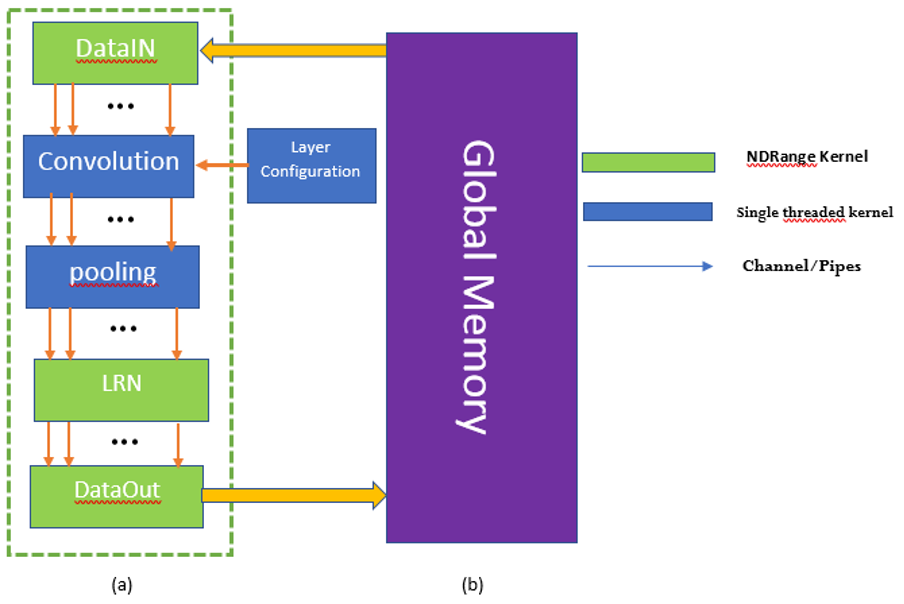} 
	\caption{Proposed CNN accelerator architecture. 
	LRN: local response normalization}
	\label{proposed}
\end{figure}

The figure\ref{proposed} illustrates the proposed architecture that consists of four kernels which are connected using Altera OpenCL extension channel/pipes.

The single threaded Convolution kernel is designed to implement both the 3D multiply-accumulate operation, defined by:
\begin{equation}
\label{eq1}
    D_0(f_0,y,x)=\sum_{f_i=1}^{C_l}\sum_{k_y=0}^{K-1}\sum_{k_x=0}^{K-1}W_l(f_0,f_i,k_y,k_x)D_i(f_i,y+k_y,x+k_x)
\end{equation}
where $D_i (f_i, y, x)$ and $D_0 (f_0, y, x)$ denote the neurons located at position $(x, y)$ in the input feature map $f_i$, and the output feature map $f_0$, respectively. $W_l (f_0, f_i, y, x)$ represents the corresponding weights in the $l^{th}$ layer which is convoluted with $f_i$. The size of the convolution filters is $K \times K$, while the total number of input feature maps is $C_l$.
In this paper, we propose to implement \ref{eq1} using a 1-D convolution structure that flattens 3-D convolution as follows:
\begin{equation}
    D_0(f_0)=\sum_{x_i=1}^{C_l\times K\times K}W_l(f_0,x_i)D_i(x_i)
\end{equation}
where $x_i$ is the index of the parameters of the layer $i$. 
 Local response normalization (LRN) layers that perform normalization operations on each inputv neuron value by a factor that depends on the neighboring neurons are also used following the pooling layer.

Therefore, we avoid nested 5-way loops levels and we get a 2-level nested loop structure, therefore, the multiplier-adder tree structure with a buffer can be efficiently pipelined by the OpenCL compiler.

Two DataIN and DataOut data transfer kernels inspired by the work of \cite{11}, two NDRange 3-D multi-mode transfer data of characteristics and weights from / to the global memory.

In addition to the most compute-intensive convolution kernel, we have designed new OpenCL kernels to speed-up layer operations widely used in CNNs, such as pooling, etc. Therefore, our proposed model can handle the CNN Forward compute stream with very small host CPU involvement, resulting in high throughput and low latency.

Cascading kernels form a deep compute pipeline able to implement a series of basic CNN operations without the need to store the interlayer data in global memory. It greatly reduces the bandwidth requirements.

\section{Results and discussions}

In this section, we present the results of the implementation of the OpenCL model proposed on the Alaric board based on Altera Arria 10 GX FPGA and the NallaTech board based on statix - 10 GX 2800.

The Arria 10 FPGA includes 660K logical elements (LE) 1687 DSP blocks and 42MB M20K, while the stratix 10 FPGA includes 2753K logical elements (LE), 5760 DSP blocks and 229MB M20K memory.

It should be noted that the card has a 2 GB DDR3 DRAM connected to the FPGA which functions as global memory for Alaric and 32 GB of DDR4 for Nallatech. OpenCL kernel codes are compiled using Altera OpenCL SDK v16.0 (Alaric) and v18.0 (Nallatech).

The host computer is equipped with an Intel Core i5-4590 processor and is running Ubuntu Linux 14.04.3. We followed the same methodology described in [11].

 and we implemented the basic design on the same Arria 10 platform. We also use the Caffe [6] convolutional learning framework as a baseline for our CPU. We extract the input image, pre-trained weights and output functions of Caffe. We compare the result of our implementation with the result of Caffe to verify functional correctness.

Two large-scale CNN models: AlexNet (8 layers) and ResNet-50 (50 layers) models were used as benchmarks to measure performance.

Since CNNs are intensive floating multiplications, the number of DSPs consumed is used as a metric for evaluating performance.
As in \cite{11} the proposed CNN design implements full-precision direct computation (32-bit float format), which also makes it favorable for implementing backpropagation flow in the learning phase of the model.
To make fair comparison, we provided the normalized performance as
”performance density” in the table. It can be noticed that the proposed implementation takes efficiently profit from the DSPs. The classification time is also better than all other implementations.

\begin{table}
	\centering
	    \caption{Comparison with other works. 2016a is in \cite{12}, FPGA2015 is in \cite{13}, and FPGRA2016b is in \cite{11} }
		\includegraphics[height=2.5in, width=4in]{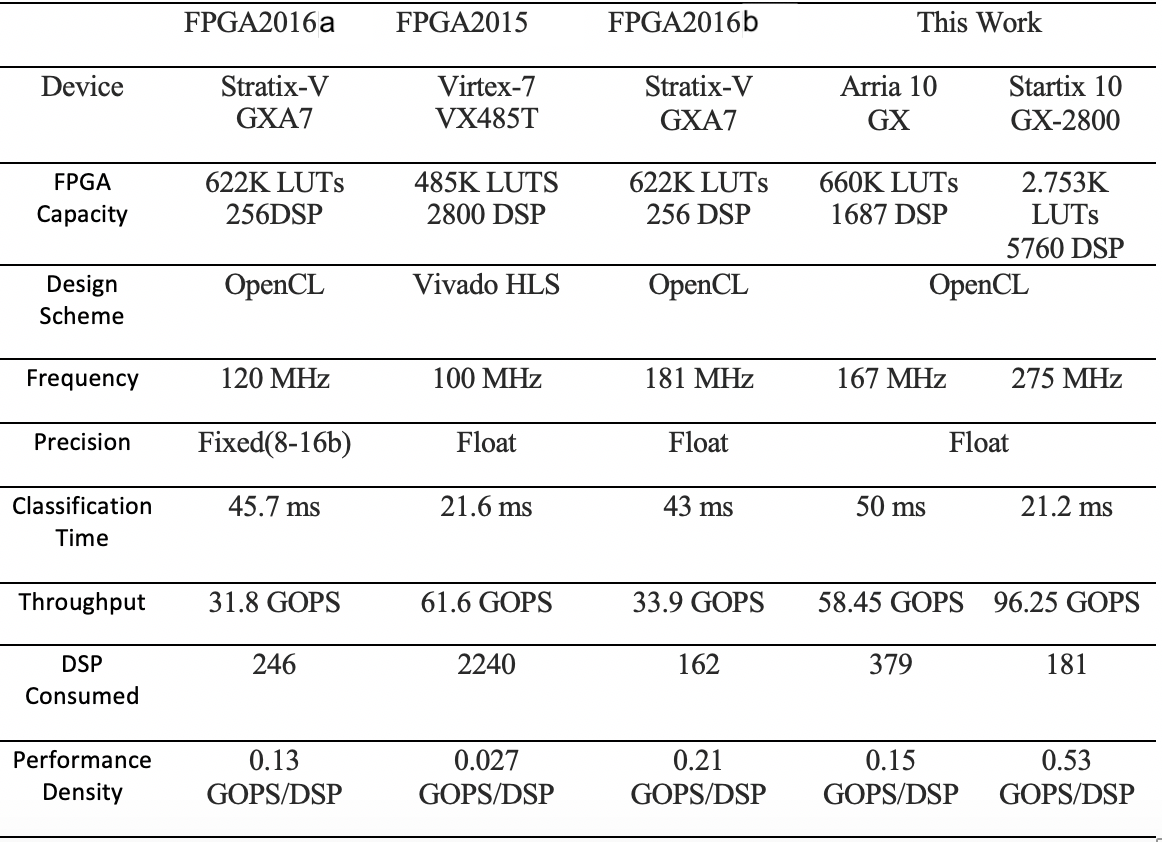} 
	
	\label{tab1}
\end{table}

\end{document}